%% file: uai_verification.tex
\documentclass[letterpaper]{article}
\usepackage{uai_2018}
\usepackage[margin=1in]{geometry}

\usepackage{times}
\usepackage{verbatim}

\title{A Dual Approach to Scalable Verification of Deep Networks}

\author{ {\bf Krishnamurthy (Dj) Dvijotham\thanks{dvij@cs.washington.edu}, Robert Stanforth, Sven Gowal, Timothy Mann, Pushmeet Kohli} \\
DeepMind \\
London, N1C 4AG, UK
}
\usepackage{color}
\usepackage{natbib}
\usepackage{graphicx}
\usepackage{amsmath}
\usepackage{amsthm}
\usepackage{mathtools}
\usepackage{amssymb}
\usepackage{subcaption}
\usepackage{color}
\usepackage{url}
\usepackage{times}
\DeclareMathOperator\arctanh{arctanh}

\newcommand{\br}[1]{\left({#1}\right)}

\newcommand{\norm}[1]{\left\| {#1} \right\|}
\DeclareMathOperator*{\argmax}{argmax}
\DeclareMathOperator*{\diag}{diag}
\newcommand{\xx}{x^{in}}
\newcommand{\xin}{x^{in}}
\newcommand{\xnom}{x^{nom}}

\newcommand{\pre}[1]{z^{#1}}
\newcommand{\preu}[1]{{\overline{z}}^{#1}}
\newcommand{\prel}[1]{{\underline{z}}^{{#1}}}
\newcommand{\post}[1]{{x}^{#1}}
\newcommand{\postu}[1]{{\overline{x}}^{{#1}}}
\newcommand{\lam}[1]{{\lambda}^{{#1}}}
\newcommand{\mum}[1]{{\mu}^{{#1}}}
\newcommand{\postl}[1]{{\underline{x}}^{{#1}}}
\newcommand{\tf}[1]{h^{#1}}
\newcommand{\num}[1]{n_{#1}}
\newcommand{\tran}[1]{{\left({#1}\right)}^T}
\newcommand{\trann}[1]{{#1}^T}
\newcommand{\Sout}{\mathcal{S}_{out}}
\newcommand{\Sin}{\mathcal{S}_{in}}
\newcommand{\pos}[1]{{\left[{#1}\right]}_+}
\newcommand{\nega}[1]{{\left[{#1}\right]}_-}
\newcommand{\tr}[1]{\mathrm{tr}\br{#1}}
\newcommand{\znom}{z^{nom}}
\newcommand{\weight}[1]{W^{#1}}
\newcommand{\bias}[1]{b^{#1}}
\newcommand{\tfs}{h}

\newtheorem{theorem}{Theorem}
\begin{document}

\maketitle

\input{intro.tex}

\input{PartA.tex}

\input{PartB.tex}

\bibliographystyle{abbrvnat}
\bibliography{references}
\newpage
\input{Appendix.tex}

\end{document}

%% file: intro.tex
\begin{abstract}
This paper addresses the problem of formally verifying desirable properties of neural networks, \emph{i.e.}, obtaining provable guarantees that neural networks satisfy specifications relating their inputs and outputs (robustness to bounded norm adversarial perturbations, for example). Most previous work on this topic was limited in its applicability by the size of the network, network architecture and the complexity of properties to be verified. In contrast, our framework applies to a general class of activation functions and specifications on neural network inputs and outputs. We formulate  verification as an optimization problem (seeking to find the largest violation of the specification) and solve a Lagrangian relaxation of the optimization problem to obtain an upper bound on the worst case violation of the specification being verified. Our approach is \emph{anytime} i.e.~it can be stopped at any time and a valid bound on the maximum violation can be obtained. We develop specialized verification algorithms with  provable tightness guarantees under special assumptions and demonstrate the practical significance of our general verification approach on a variety of verification tasks.
\end{abstract}

\section{INTRODUCTION}
Neural networks and deep learning have revolutionized machine learning achieving state of the art performance on a wide range of complex prediction tasks \citep{krizhevsky2012imagenet,Goodfellow-et-al-2016}. However, in recent years, researchers have observed that even state of the art networks can be easily fooled into changing their predictions by making small but carefully chosen modifications to the input data (known as 
\emph{adversarial perturbations}) \citep{szegedy2013intriguing, kurakin2016adversarial,carlini2017adversarial, goodfellow2014explaining, carlini2017towards}. While modifications to neural network training algorithms have been proposed to mitigate this phenomenon\cite{madry2018towards}, a comprehensive solution that is fully robust to adversarial attacks remains elusive~\citep{carlini2017towards, uesato2018adversarial}.  

Neural networks are typically tested using the standard machine learning paradigm: If the performance (accuracy) of the network is sufficiently high on a holdout (test) set that the network did not have access to while training, the network is deemed acceptable. This is justified by statistical arguments based on an \emph{i.i.d.~}assumption on the data generating mechanism, that is each input output pair is generated independently from the same (unknown) data distribution. However, this evaluation protocol is not sufficient in domains with critical safety constraints \citep{marston2015acas}. In these cases, we may require a stronger test: for example, we may require that the network is robust against adversarial perturbations within certain bounds.

\paragraph{Adversarial evaluation.}  In the context of adversarial examples, a natural idea is to test neural networks by checking if it is possible to generate an adversarial attack to change the label predicted by the neural network \citep{kurakin2016adversarial} and train them to be robust to these examples \cite{madry2018towards}. Generating adversarial examples is a challenging computational task itself, and the attack generated by a specific attack algorithm may be far from optimal. This may lead one to falsely conclude that a given model is robust to attacks even though a stronger adversary may have broken the robustness. Recent work \citep{athalye2018obfuscated, uesato2018adversarial} has shown that evaluating models against weak adversaries can lead to incorrect conclusions regarding the robustness of the model. Thus, there is a need to go beyond evaluation using specific adversarial attacks and find approaches that provide provable guarantees against attacks by \emph{any adversary}.

\paragraph{Towards verifiable models.}
Verification of neural networks has seen significant research interest in recent years. In the formal verification community, Satisfiability Modulo Theory (SMT) solvers have been adapted for verification of neural networks~\citep{ElhersPlanet,huang2017safety,Reluplex}. While SMT solvers have been successfully applied to several domains, applying them to large neural networks remains a challenge due to the scale of the resulting SMT problem instances. Furthermore, these approaches have been largely limited to networks with piecewise linear activation functions since most SMT solvers are unable to deal efficiently with nonlinear arithmetic.
More recently, researchers have proposed a set of approaches that make use of branch and bound algorithms either directly or via mixed-integer programming solvers
\citep{bunel2017piecewise,
cheng2017maximum,tjeng2017verifying}. While these approaches achieve strong results on smaller networks, scaling them to large networks remains an open challenge. These approaches also rely heavily on the piecewise linear structure of networks where the only nonlinearities are maxpooling and ReLUs.

\paragraph{Towards scalable verification of general models.}
In this paper, we develop a novel approach to neural network verification based on optimization and duality. The approach consists of formulating the verification problem as an optimization problem that tries to find the largest violation of the property being verified. If the largest violation is smaller than zero, we can conclude that the property being verified is true. By using ideas from duality in optimization, we can obtain bounds on the optimal value of this problem in a computationally tractable manner. Note that this approach is \emph{ sound but incomplete}, in that there may be cases where the property of interest is true, but the bound computed by our algorithm is not tight enough to prove the property.
This strategy has been used in prior work as well \citep{kolter2017provable, CertifiedDefenses}. However, our results improve upon prior work in the following ways:
\begin{enumerate}
    \item Our verification approach applies to \emph{arbitrary} feedforward neural networks with \emph{any architecture} and \emph{any activation function} and our framework recovers previous results \citep{ElhersPlanet} when applied to the special case of piecewise linear activation functions.
    \item We can handle verification of systems with discrete inputs and combinatorial constraints on the input space, including cardinality constraints.
    \item The computation involved only requires solving an unconstrained convex optimization problem (of size linear in the number of neurons in the network), which can be done using a subgradient method efficiently. Further, our approach is \emph{anytime}, in the sense that the computation can be stopped at any time and a valid bound on the verification objective can be obtained. 
    \item For the special case of single hidden layer networks, we develop specialized verification algorithms with provable tightness guarantees.
    \item We attain state of the art verified bounds on adversarial error rates on image classifiers trained on MNIST and CIFAR-10 under adversarial perturbations in the infinity norm. 
\end{enumerate}

\section{Related Work}

\paragraph{Certifiable training and verification of neural networks}
A separate but related thread of work is on \emph{certifiable training}, ie, training neural networks so that they are guaranteed to satisfy a desired property (for example, robustness to adversarial examples within a certain radius) \citep{kolter2017provable,CertifiedDefenses}. These approaches use ideas from convex optimization and duality to construct bounds on an optimization formulation of  verification. However, these approaches are limited to either a class of activation functions (piecewise linear models) or architectures (single hidden layer, as in \cite{CertifiedDefenses}). Further, in \citet{kolter2017provable}, the dual problem starts with a constrained convex formulation but is then converted into an unconstrained but nonconvex optimization problem to allow for easy optimization via a backprop-style algorithm. In contrast, our formulation allows for an unconstrained dual convex optimization problem so that for \emph{any choice of dual variables}, we obtain a valid bound on the adversarial objective and this dual problem can be solved efficiently using subgradient methods. 

We also note that the ultimate goals of \citep{kolter2017provable, CertifiedDefenses} are different from our paper: they modify the training procedure of the neural network so that the network is trained to be easily verifiable. In contrast, our work focuses on extending verification algorithms to apply to a broader class of architectures, activation functions and - in this sense, we view our work as complementary to \citep{kolter2017provable,CertifiedDefenses}. In fact, since the objective of our dual optimization is differentiable with respect to the network weights, we can extend our approach to training verifiable networks easily by simultaneously optimizing the network weights and dual variables to minimize the dual objective. We leave the study of such an extension for future work.

\paragraph{Theoretical analysis of robustness}
Another related line of work has to do with theoretical analysis of adversarial examples. It has been shown that feedforward ReLU networks cannot learn to distinguish between points on two concentric spheres without necessarily being vulnerable to adversarial examples within a small radius \citep{Adv_spheres}. Under a different set of assumptions, the existence of adversarial examples with high probability is also established in \citet{fawzi2018adversarial}. In \citet{wang2017analyzing}, the authors study robustness of nearest neighbor classifiers to adversarial examples. As opposed to these theoretical analyses, our approach searches computationally for proofs of existence or non-existence of adversarial examples. The approach does not say anything a-priori about the existence of adversarial examples, but can be used to investigate their existence for a given network and compare strategies to guard against adversarial attacks.

%% file: PartA.tex
\section{VERIFICATION AS OPTIMIZATION}
\subsection{NOTATION}

Our techniques apply to general feedforward architectures and recurrent networks, but we focus on layered architectures for the development in this paper. The input layer is numbered $0$, the hidden layers are numbered $1, \ldots, L-1$ and the output layer is numbered $L$. The size of layer $l$ is denoted $
\num{l}$

We denote by $\xx$ the input to the neural network, by $\pre{l}$ the pre-activations of neurons at layer $l$ \emph{before} application of the activation function and by $\post{l}$ the vector of neural activations \emph{after} application of the activation function (to $\pre{l-1}$). For convenience, we define $\post{0}=\xx$. We use $\post{l}(\xx), \pre{l}(\xx)$ to denote the activations at the $l$-th layer as a function of the input $\xx$. Upper and lower bounds on the pre/post activations are denoted by $\postl{l}, \postu{l}, \prel{l}, \preu{l}$ respectively. The activation function at layer $l$ is denote $\tf{l}$ and is assumed to be applied component-wise, ie,
\[[\tf{l}(\pre{l})]_k = \tf{l}_k(\pre{l}_k)\]
Note that max-pooling is an exception to this rule - we discuss how max-pooling is handled separately in the Appendix section \ref{sec:AppMaxPool}. 
The weights of the network at layer $l$ are denoted $\weight{l}$ and the bias is denoted $\bias{l}$, $\pre{l}=\weight{l}\post{l}+\bias{l}$. 

\subsection{VERIFICATION PROBLEM}
As mentioned earlier, \emph{verification } refers to the process of checking that the output of the neural network satisfies a certain desirable property for all choices of the input within a certain set. Formally, this can be stated as follows:
\begin{align}
\forall \xx \in \Sin\br{\xnom} \quad \post{L}(\xx) \in \Sout \label{eq:Verify}
\end{align}
where $\xx$ denotes the input to the network, $\xnom$ denotes a nominal input, $\Sin(\xnom)$ defines the constrained subset of inputs induced by the nominal input, and $\Sout$ denotes the constraints on the output that we would like to verify are true for all inputs in $\Sin(\xnom)$. In the case of adversarial perturbations in image classification, $\xnom$ would refer to the nominal (unperturbed image),  $\Sin(\xnom)$ would refer to all the images that can be obtained by adding bounded perturbations to $\xnom$, and $\xin$ would refer to a perturbed image. 

In this paper, we will assume that: $\Sout$ is always a described by a finite set of linear constraints on the values of final layer {\em ie.} $\Sout =\cap_{i=1}^m \{\post{L}: \tran{c^{i}} \post{L} + d^i \leq 0\}$, and 
$\Sin\br{\xnom}$ is any bounded set such that any linear optimization problem of the form 
\[\max_{\xx\in \Sin\br{\xnom}} c^Tx\]
can be solved efficiently. This includes convex sets and also sets describing combinatorial structures like spanning trees, cuts in a graph and cardinality constraints.

See the following examples for a concrete illustration of the formulation of the problem: \\
\paragraph{Robustness to targeted adversarial attacks.} Consider an adversarial attack that seeks to perturb an input $\xnom$ to an input $\xx$ subject to a constraint on the perturbation $\norm{\xx-\xnom} \leq \epsilon$ to change the label from the true label $i$ to a target label $j$. We can map this to \eqref{eq:Verify} as follows: 
\begin{subequations}
\begin{align}
& \Sin\br{\xnom}=\{\xx: \norm{\xx-\xnom} \leq \epsilon\}, \\
& \Sout=\{z: c^Tz \leq 0\}    
\end{align}
\end{subequations}
where $c$ is a vector with $c_j=1, c_i=-1$ and all other components $0$. Thus, $\Sout$ denotes the set of outputs for which the true label $i$ has a higher logit value than the target label $j$ (implying that the targeted adversarial attack did not succeed). \\
\paragraph{Monotonic predictors.} Consider a network with a single real valued output and we are interested in ensuring that the output is monotonically increasing wrt each dimension of the input $\xx$. We can state this as a verification problem:
\begin{subequations}
\begin{align}
& \Sin\br{\xnom}=\{\xin: \xin \geq \xnom\} \\
& \Sout=\{\post{L}:  \post{L}\br{\xnom}-\post{L} \leq 0\}  
\end{align} 
\end{subequations}
Thus, $\Sout$ denotes the set of outputs which are large than the network output at $\xnom$. If this is true for each value of $\xnom$, then the network is monotone.

\paragraph{Cardinality constraints.} In several cases, it makes sense to constrain a perturbation not just in norm but also in terms of the number of dimensions of the input that can be perturbed. We can state this as:
\begin{subequations}
\begin{align}
& \Sin\br{\xnom}=\nonumber \\
&\, \{\xin: \norm{\xin-\xnom}_0 \leq k, \norm{\xin-\xnom}_\infty \leq \epsilon\} \\
& \Sout=\{z: c^Tz \leq 0\}    
\end{align} 
where $\norm{x}_0$ denotes the number of non-zero entries in x.
\end{subequations}
Thus, $\Sout$ denotes the set of outputs which are larger than the network output at $\xnom$. If this is true for each value of $\xnom$, then the network is monotone.

\subsection{OPTIMIZATION PROBLEM FOR VERIFICATION}\label{sec:MainDual}
Once we have a verification problem formulated in the form \eqref{eq:Verify}, we can easily turn the verification procedure into an optimization problem. This is similar to the optimization based search for adversarial examples \citep{szegedy2013intriguing} when the property being verified is adversarial robustness. For brevity, we only consider the case where $\Sout$ is defined by a single linear constraint $c^Tz + d \leq 0$. If there are multiple constraints, each one can be verified separately.
\begin{subequations}
\begin{align}
& \max_{\substack{\pre{0}, \ldots, \pre{L-1}\\ \post{0}, \ldots, \post{L}}}  c^T \post{L} + d \\
& \text{s.t }  \post{l+1} = \tf{l}\br{\pre{l}}, l = 0, 1, \ldots, L-1 \label{eq:RobustVerifya}\\
& \pre{l} = \weight{l}\post{l} + \bias{l}, l = 0, 1, \ldots, L-1 \label{eq:RobustVerifyb}\\
& \post{0} = \xx, \xx \in \Sin\br{\xnom} \label{eq:RobustVerifyc}
\end{align}\label{eq:RobustVerify}
\end{subequations}
If the optimal value of this problem is smaller than $0$ (for each $c,d$ in the set of linear constraints defining $\Sout$), we have verified the property \eqref{eq:Verify}.  This is a nonconvex optimization problem and finding the global optimum in general is NP-hard (see Appendix section \ref{sec:NP-hard} for a proof). However, if we can compute \emph{upper bounds} on the value of the optimization problem and the upper bound is smaller than $0$, we have successfully verified the property. In the following section, we describe our main approach for computing bounds on the optimal value of \eqref{eq:RobustVerify}.

\subsection{BOUNDING THE VALUE OF THE OPTIMIZATION PROBLEM}
We assume that bounds on the activations $\pre{l},\post{l}, l=0, \ldots, L-1$ are available. Section \ref{sec:BoundTighten} discusses details of how such bounds may be obtained given the constraints on the input layer $\Sin\br{\xnom}$. We can bound the optimal value of \eqref{eq:RobustVerify} using a Lagrangian relaxation of the constraints:
\begin{subequations}
\begin{align}
 \max_{\substack{\pre{0}, \ldots, \pre{L-1}\\ \post{0}, \post{1}, \ldots, \post{L-1}}} &  c^T\br{ \tf{L-1}\br{\pre{L-1}}} + d \nonumber\\
& \, + \sum_{l=0}^{L-1} \tran{\mum{l}}\br{\pre{l}-\weight{l}\post{l}-\bias{l}} \nonumber\\
& \, + \sum_{l=0}^{L-2} \tran{\lam{l}}\br{\post{l+1}-\tf{l}\br{\pre{l}}} \\
\text{s.t. } & \prel{l} \leq \pre{l} \leq  \preu{l} \quad  ,l = 0, 1, \ldots, L-1  \\
& \postl{l} \leq \post{l} \leq  \postu{l} \quad  ,l = 0, 1, \ldots, L-1 \\
& \post{0} \in \Sin\br{\xnom}
\end{align} \label{eq:DualObj}
\end{subequations}
Note that any feasible solution for the original problem \eqref{eq:RobustVerify} is feasible for the above problem, and for any such solution, the terms involving $\lambda, \mu$ become $0$ (since the terms multiplying $\lambda, \mu$ are $0$ for every feasible solution). Thus, for any choice of $\lambda, \mu$, the above optimization problem provides a valid upper bound on the optimal value of \eqref{eq:RobustVerify} (this property is known as weak duality \citep{vandenberghe2004convex}).

We now look at solving the above optimization problem. Since the objective and constraints are separable in the layers, the variables in each layer can be optimized independently. For $l=1, \ldots, L-1$, we have
\begin{align*}
& f_l\br{\lam{l-1}, \mum{l}} = \\
& \max_{\post{l} \in [\postl{l}, \postu{l}]} \tran{\lam{l-1}-\tran{\weight{l}}\mum{l}}\post{l} - \tran{\bias{l}}\mum{l}
\end{align*}
which can be solved trivially by setting each component of $\post{l}$ to its upper or lower bound depending on whether the corresponding entry in $\lam{l-1}-\tran{\weight{l}}\mum{l}$ is non-negative. Thus,
\begin{align*}
& f_l\br{\lam{l-1}, \mum{l}} = \\
&  \trann{\pos{\lam{l-1}-\tran{\weight{l}}\mum{l}}}\postu{l}  \\
& \, + \trann{\nega{\lam{l-1}-\tran{\weight{l}}\mum{l}}}\postl{l}- \tran{\bias{l}}\mum{l}
\end{align*}
where $\pos{x}=\max(x, 0), \nega{x}=\min(x, 0)$ denote the positive and negative parts of $x$.

Similarly, collecting the terms involving $\pre{l}$, we have, for $l=0, \ldots, L-1$
\[\tilde{f}_l(\lam{l}, \mum{l}) = \max_{\pre{l} \in [\prel{l}, \preu{l}]} \trann{\mum{l}}\pre{l} -\tran{\lam{l}}\tf{l}\br{\pre{l}} \]
where $\lambda_{L-1} = -c$.

Since $\tf{l}$ is a component-wise nonlinearity, each dimension of $\pre{l}$ can be optimized independently. For the $k$-th dimension, we obtain
\[\tilde{f}_{l, k}\br{\lam{l}_k, \mum{l}_k} = \max_{\pre{l}_k \in [\prel{l}_k, \preu{l}_k]} \mum{l}_k \pre{l}_k - \lam{l}_k \tf{l}_k\br{\pre{l}_k} \]
This is a one-dimensional optimization problem and can be solved easily- for common activation functions (ReLU, tanh, sigmoid, maxpool), it can be solved analytically, as discussed in appendix section \ref{sec:AppA}. 
Finally, we need to solve
\[f_0(\mum{0}) = \max_{\post{0} \in \Sin\br{\xnom}} \tran{-\tran{\weight{0}}\mu_0}\post{0} - \tran{\bias{0}}\mum{0}\]
which can also be solved easily given the assumption on $\Sin$. We work out some concrete cases in \ref{sec:AppB}.

Once these problems are solved, we can construct the dual optimization problem:
\begin{align}
 \min_{\substack{\lambda, \mu}} & \sum_{k=0}^{\num{L-1}} \tilde{f}_{L-1, k}\br{-c_k, \mum{l}_k} +  \sum_{l=0}^{L-2}  \sum_{k=0}^{\num{l}} \tilde{f}_{l, k}\br{\lam{l}_k, \mum{l}_k}\nonumber\\
 & + \sum_{l=1}^{L-1}  f_l(\lam{l-1}, \mum{l})+ f_0\br{\mum{0}}+d\label{eq:DualOpt}
\end{align}
This seeks to choose the values of $\lambda, \mu$ so as to minimize the upper bound on the verification objective, thereby obtaining the tightest bound on the verification objective.

This optimization can be solved using a subgradient method on $\lambda, \mu$. 

\begin{theorem}\label{thm:Main}
For any values of $\lambda, 
\mu$, the objective of \eqref{eq:DualOpt} is an upper bound on the optimal value of \eqref{eq:RobustVerify}. Hence, the optimal value of \eqref{eq:DualOpt} is also an upper bound. Further, \eqref{eq:DualOpt} is a convex optimization problem in $\br{\lambda, \mu}$.
\end{theorem}
\begin{proof}
The upper bound property follows from weak duality \citep{vandenberghe2004convex}. The fact that \eqref{eq:DualOpt} is a convex optimization problem can be seen as each term $f_l, \tilde{f}_{l, k}$ is expressed as a maximum overa set of linear functions of $\lambda, \mu$ \citep{vandenberghe2004convex}.
\end{proof}

\begin{theorem}\label{thm:Planet}
If each $h$ is a ReLU function, then \eqref{eq:DualOpt} is equivalent to the dual of the LP described in \citet{ElhersPlanet}.
\end{theorem}
\begin{proof}
See section \ref{sec:TheoryRes}.
\end{proof}
The LP formulation from \citet{ElhersPlanet} is also used in \citet{kolter2017provable}. The dual of the LP is derived in \citet{kolter2017provable} - however this dual is different from \eqref{eq:DualOpt} and ends up with a constrained optimization formulation for the dual (the details of this can be found in appendix section \ref{sec:AppCompareRelax}). To allow for an unconstrained formulation, this dual LP is transformed to a backpropagation-like computation. While this allows for folding the verification into training, it also introduces nonconvexity in the verification optimization - our formulation of the dual differs from \citet{kolter2017provable} in that we directly solve an unconstrained dual formulation, allowing us to circumvent the need to solve a non-convex optimization for verification.



%% file: PartB.tex
\subsection{TOWARDS THEORETICAL GUARANTEES FOR VERIFICATION}

The bounds computed by solving \eqref{eq:DualOpt} could be loose in general, since \eqref{eq:RobustVerify} is an NP-hard optimization problem (section \ref{sec:NP-hard}). SMT solvers and MIP solvers are guaranteed to find the exact optimum for piecewise linear neural networks, however, they may take exponential time to do so. Thus, an open question remains: Are there cases where it is possible to perform \emph{exact verification} efficiently? If not, can we approximate the verification objective to within a certain factor (known a-priori)? We develop results answering these questions in the following sections.

\emph{Prior work:}  For any linear classifier, the scores of each label are a linear function of the inputs $w^T_i x+b_i$. Thus, the difference between the predictions of two classes $j$ (target class for an adversary) and class $i$ (true label) is $\tran{w_i-w_j}x$. Maximizing this subject to $\norm{x-x^0}_2\leq \epsilon$ can be solved analytically to obtain the value
$\tran{w_i-w_j}x^0 + \norm{w_i-w_j}_2\epsilon$. This observation formed the basis for algorithms in \citep{CertifiedDefenses} and \citep{hein2017formal}. However, once we move to nonlinear classifiers, the situation is not so simple and computing the worst case adversarial example, even in the 2-norm case, becomes a challenging task. In \citep{hein2017formal}, the special case of kernel methods and single hidden layer classifiers are considered, but the approaches developed are only upper bounds on the verification objective (just like those computed by our dual relaxation approach). Similarly, in \citet{CertifiedDefenses}, a semidefinite programming approach is developed to compute bounds on the verification objective for the special case of adversarial perturbations on the infinity norm. However, none of these approaches come with a-priori guarantees on the quality of the bound, that is, before actually running the verification algorithm, one cannot predict how tight the bound on the verification objective would be. In this section, we develop novel theoretical results that quantify when the verification problem \eqref{eq:RobustVerify} can be solved \emph{either exactly or with a-priori approximation guarantees}. Our results require strong assumptions and do not immediately apply to most practical situations. However, we believe that they shed some understanding on the conditions under which exact verification can be performed tractably and lead to specialized verification algorithms that merit further study.

We assume the following for all results in this section:\\
1) We study networks with a single hidden layer,  \emph{i.e.} $L=2$, with activation function $\tf{0}=\tfs$ and a linear mapping from the penultimate to the output layer $\post{2}=\tf{1}\br{\pre{1}}=\pre{1}=\weight{1}\post{1}+\bias{1}$. \\
2) The network has a differentiable activation function $h$ with Lipschitz-continuous derivatives denoted $h^{\prime}$ (tanh, sigmoid, ELU, polynomials satisfy this requirement).\\
3) $\Sin\br{\xnom}= \{\xin: \norm{\xin-\xnom}_2 \leq \epsilon\}$.

Since the output layer is a linear function of the penultimate layer $\post{1}$, we have 
\begin{align*}
& c^T\post{2}= \tran{\tran{\weight{1}}c}\post{1}+c^T\bias{1}\\
&=\tran{\tran{\weight{1}}c}\tf{0}\br{\pre{0}}+c^T\bias{1}    
\end{align*}
For brevity, we simply denote $\tran{\weight{1}}c$ as $c$, drop the constant term $c^T\bias{1}$ and let $W=\weight{0},b=\bias{0}, \znom=W\xnom+b$ and $W_i$ denote the $i$-th row of $W$. Then, \eqref{eq:RobustVerify} reduces to:
\begin{align}
\max_{\xin:\norm{\xin-\xnom}_2 \leq \epsilon} & \sum_i c_i \tfs_i\br{W_i\xin + b_i} \label{eq:DualSpl}
\end{align} 
\begin{theorem}\label{thm:Exact}
Suppose that $h$ has a Lipschitz continuous first derivative:
\[h^{\prime}_i(t)-h^{\prime}_i(\tilde{t})\leq \gamma_i |t-\tilde{t}|\]
Let 
\begin{align*}
& \nu = \norm{\diag\br{c}W^T h^\prime\br{z^{nom}}}_2\\
& L = \sigma_{\max}\br{\diag\br{c}W^T}\sigma_{\max}\br{\diag\br{\gamma}W}
\end{align*}
Then $\forall$ $\epsilon \in (0, \frac{\nu}{2L})$, the iteration:
\[x^{k+1} \gets \xnom + \epsilon\br{\frac{W^T\diag\br{c} h^\prime(Wx^k+b)}{\norm{W^T\diag\br{c} h^\prime(Wx^k+b)}_2}}\]
starting at $x^0=\xnom$
converges to the global optimum $x^\star$ of \eqref{eq:DualSpl} at the rate $\norm{x^k-x^\star} \leq {\br{\frac{\epsilon L}{\nu-\epsilon L}}}^k$
\end{theorem}
\begin{proof}
Section \ref{sec:TheoryRes}
\end{proof}

Thus, when $\epsilon$ is small enough, a simple algorithm exists to find the global optimum of the verification objective. However, even when $\epsilon$ is larger, one can obtain a good approximation of the verification objective. In order to do this, consider the following quadratic approxiimation of the objective from  \eqref{eq:DualSpl}:
\begin{subequations}
\begin{align}
 \max  & \sum_i c_i\br{ \tfs_i\br{\znom_i}+\tfs_i^\prime\br{\znom_i}\br{W_iz}} \nonumber\\
& \quad + \sum_i \frac{c_i}{2}\tfs_i^{\prime\prime}\br{\znom_i}{\br{W_iz}}^2\\
& \text{s.t. }  \norm{z}_2 \leq \epsilon
\end{align}\label{eq:DualSplApprox}
\end{subequations}
This optimization problem corresponds to a trust region problem that can be solved to global optimality using semidefinite programming \citep{yakubovic1971s}:
\begin{subequations}
\begin{align}
\max_{z, Z}  & \sum_i c_i\br{ \tfs_i\br{\znom_i}+\tfs_i^\prime\br{\znom_i}(W_iz)} \nonumber \\
& \, + \sum_i \frac{c_i}{2}\tfs_i^{\prime\prime}\br{\znom_i}\tr{W_i^TW_iZ}\\
\text{s.t. } &  \tr{Z}\leq \epsilon,  \begin{pmatrix}
1 & z^T \\
z & Z
\end{pmatrix} \succeq 0
\end{align}\label{eq:DualSplApproxSDP}
\end{subequations}
where $X \succeq 0$ denotes that $X$ is constrained to be a positive semidefinite matrix. 
While this can be solved using general semidefinite programming solvers, several special purpose algorithms exist for this trust region problem that can exploit its particular structure for efficient solution, \citep{hazan2016linear}

\begin{theorem}\label{thm:Approx}
Suppose that $h$ is thrice-differentiable with a globally bounded third derivative. 
Let 
\begin{align*}
& \zeta_i = \norm{W_i}_2, \eta_i =  \sup_{t} |h^{\prime\prime\prime}_i(t)|,\kappa = \frac{1}{6}\br{\sum_i \eta_i c_i \zeta_i^3}
\end{align*}
For each $\epsilon > 0$, the difference between the optimal values of \eqref{eq:DualSplApproxSDP}, \eqref{eq:DualSpl} is at most $\kappa\epsilon^3$.
\end{theorem}
\begin{proof}
See Section \ref{sec:TheoryRes}
\end{proof}

\section{EXPERIMENTS}
In this section, we present numerical studies validation our approach on three sets of verification tasks:\\
\emph{Image classification on MNIST and CIFAR:} We use our approach to obtain guaranteed lower bounds on the accuracy of image classifers trained on MNIST and CIFAR-10 under adversarial attack with varying sizes of the perturbation radius. We compare the bounds obtained by our method with prior work (in cases where prior work is applicable) and also with the best attacks found by various approaches.     \\
\emph{Classifier stability on GitHub data:} We train networks on sequences of commits on GitHub over a collection of 10K repositories - the prediction task consists of predicting whether a given repository will reach more than 40 commits within 250 days given data observed until a certain day. Input features consist a value between 0 and 1 indicating the number of days left until the 250th day, as well as another value indicating the progress of commits towards the total of 40. As the features evolve, the prediction of the classifier changes (for example, predictions should become more accurate as we move closer to the 250th day). In this situation, it is desirable that the classifier provides consistent predictions and that the number of times its prediction switches is as small as possible. It is also desirable that this switching frequency cannot be easily be changed by perturbing input features. We use our verification approach combined with dynamic programming to compute a bound on the maximum number of switches in the classifier prediction over time.\\
\emph{Digit sum task:} We consider a more complex verification task here: Given a pair of MNIST digits, the goal is to bound how much the sum of predictions of a classifier can differ from the true sum of those digits under adversarial perturbation subject to a total budget on the perturbation across the two digits.

\subsection{IMAGE CLASSIFICATION: MNIST AND CIFAR}


We study adversarial attacks subject to an $l_\infty$ bound on the input perturbation. An adversarial example (AE) is a pertubation of an input of the neural network such that the output of the neural network differs from the correct label for that input. An AE is said to be within radius $\epsilon$ if the $\ell_\infty$ norm of the difference between the AE and the original input is smaller than $\epsilon$. We are interested in the \emph{adversarial error rate}, that is,
\[\frac{\#\text{ Test examples that have an AE within radius }\epsilon}{\text{Size of test set}}\]
Computing this quantity precisely requires solving the NP-hard problem \eqref{eq:RobustVerify} for each test example, but we can obtain upper bounds on it using our (and other) verification methods and lower bounds using a fixed attack algorithm (in this paper we use a bound constrained LBFGS algorithm similar to \citep{carlini2017towards}). Since theorem \ref{thm:Planet} shows that for the special case of piecewise linear neural networks, our approach reduces to the basic LP relaxation from \citet{ElhersPlanet} (which also is the basis for the algorithms in \citet{bunel2017piecewise} and \citet{kolter2017provable}), we focus on networks with smooth nonlinearities like tanh and sigmoid. We compare our approach with The SDP formulation from \citet{CertifiedDefenses} (note that this approach only works for single hidden layer networks, so we just show it as producing vacuous bounds for other networks).

Each approach gets a budget of $300$ s per verifiation problem (choice of test example and target label). Since the SDP solver from \citep{CertifiedDefenses} only needs to be run once per label pair (and not per test example), its running time is amortized appropriately. 

\begin{figure*}[tb]
    \centering
    \begin{subfigure}[b]{0.45\textwidth}
        \includegraphics[width=\textwidth]{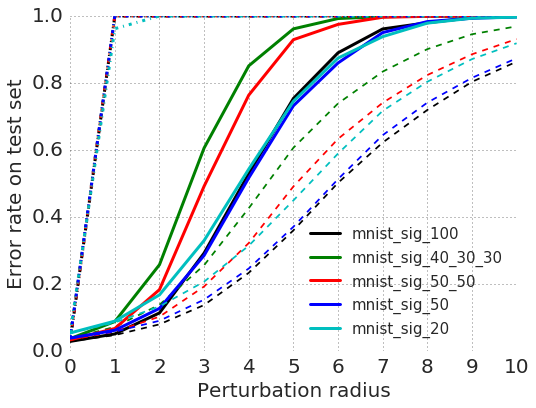}
        \caption{Sigmoid}
        \label{fig:sig}
    \end{subfigure}
    ~ 
    \begin{subfigure}[b]{0.45\textwidth}
        \includegraphics[width=\textwidth]{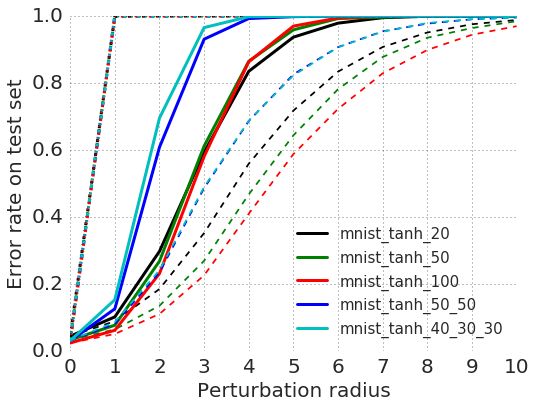}
        \caption{Tanh}
        \label{fig:tanh}
    \end{subfigure}
    ~ 
\caption{Figures  show three curves per model - Dashed line: Lower bound from LBFGS attack. Solid line: Our verified upper bound. Dash-dot line: SDP verified upper bound from \citep{CertifiedDefenses}. Each color represents a different network.  The dashed lines at the bottom are lower bounds on the error rate computed using the best attack found using the LBFGS algorithm.}\label{fig:MNIST}    
\end{figure*}

\emph{Results on smooth activation functions:} Figures \ref{fig:sig},\ref{fig:tanh} show  that our approach is able to compute nearly tight bounds (bounds that match the upper bound) for small perturbation radii (up to 2 pixel units) and our bounds significantly outperform those from the SDP approach \citep{CertifiedDefenses} ( which is only able to compute nontrivial bounds for the smallest model with 20 hidden units). 

\begin{figure}[htb]
\begin{subfigure}[b]{.49\textwidth}
        \includegraphics[width=\columnwidth]{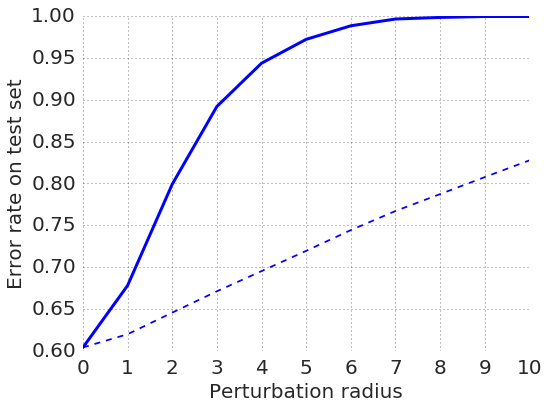}
    \caption{Verification of robust CIFAR-10 model as a function of perturbation radius $\epsilon$.}\label{fig:CIFAR}
\end{subfigure}    
\begin{subfigure}[b]{.49\textwidth}
        \includegraphics[width=\columnwidth]{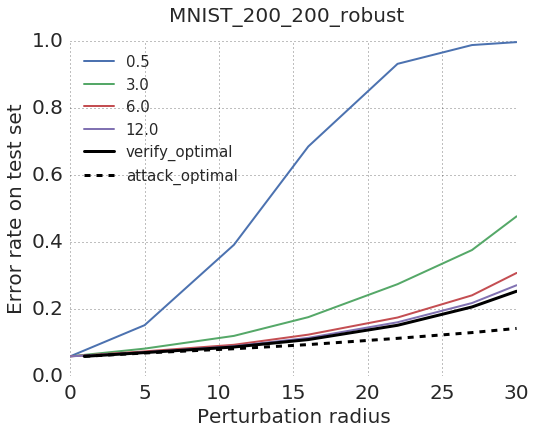}
    \caption{Verification of robust models as a function of perturbation radius $\epsilon$. Solid lines:verified upper bounds and dashed lines: attack lower bounds.}\label{fig:mnist_robust}
\end{subfigure}
\end{figure}

\begin{figure}[htb]
\begin{subfigure}[b]{.49\textwidth}
    \centering
    \includegraphics[width=\columnwidth]  {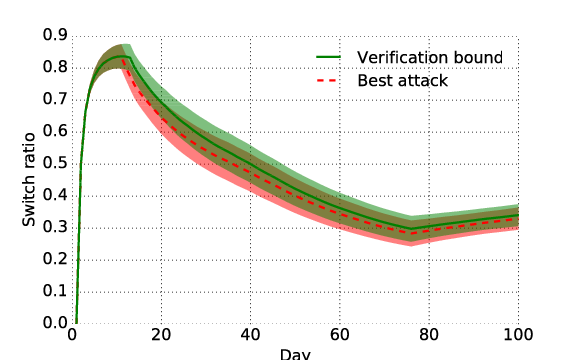}
    \caption{Maximum number of switches (over number of days evaluated so far) over different days (upper bound from our verification approach and lower bound from the best attack found). Results are averaged over 100 unseen repositories. Shaded areas are 95\% confidence intervals.}
    \label{fig:github_example}
\end{subfigure}
\begin{subfigure}[b]{.49\textwidth}
    \centering
    \includegraphics[width=\columnwidth]  {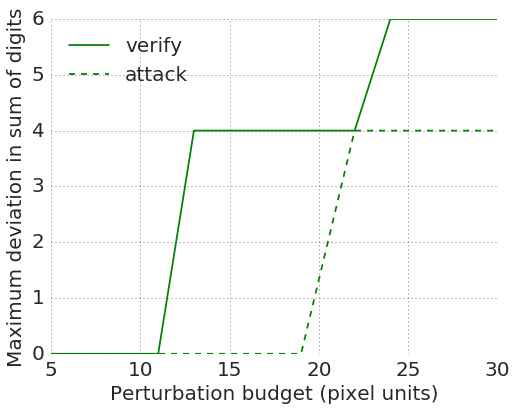}
    \caption{Maximum difference between predicted and actual sum of digits vs the perturbation budget for a pair of randomly chosen MNIST images.}
    \label{fig:complex_verify}
    \end{subfigure}
\end{figure}

\emph{Results on models trained adversarially:} We use the adversarial training approach of \citep{madry2018towards} and train models on MNIST and CIFAR that are robust to perturbations from the LBFGS-style attack on the training set. We then apply our verification algorithm to these robust models and obtain bounds on the adversarial error rate on the test set. These models are all multilayer models, so the SDP approach from \citep{CertifiedDefenses} does not apply and we do not plot it here. We simply plot the attack versus the bound from our approach. The results for MNIST are plotted in figure \ref{fig:mnist_robust} and for CIFAR in figure \ref{fig:CIFAR}. On networks trained using a different procedure, the approaches from \citet{CertifiedDefenses} and \citet{kolter2017provable} are able to achieve stronger results for larger values of $\epsilon$ (they work with $\epsilon=.1$ in real units, which corresponds to $\epsilon=26$ in pixel units we use here). However, we note that our verification procedure is agnostic to the training procedure, and can be used to obtain fairly tight bounds for any network and any training procedure. In comparison, the results in \citet{kolter2017provable} and \citet{CertifiedDefenses} rely on the training procedure optimizing the verification bound. Since we do not rely on a particular adversarial training procedure, we were also able to obtain the first non-trivial verification bounds on CIFAR-10 (to the best of our knowledge) shown in figure \ref{fig:CIFAR}. While the model quality is rather poor, the results indicate that our approach could scale to more complicated models.

\subsection{GITHUB CLASSIFIER STABILITY}


We allow the adversary to modify input features by up to 3\% (at each timestep) and our goal is to bound the maximum number of prediction switches induced by each attack over time. We can model this within our verification framework as follows: Given a sequence of input features, we compute the maximum number of switches achievable by first computing the target classes that are reachable through an adversarial attack at each timestep (using \eqref{eq:DualOpt}), and then running a dynamic program to compute the choices of target classes over time (from within the reachable target classes) to maximize the number of switches over time.

Figure~\ref{fig:github_example} shows how initially predictions are easily attackable (as little information is available to make predictions), and also shows how the gap between our approach and the best attack found using the LBFGS algorithm evolves over time.

\subsection{COMPLEX VERIFICATION TASK: DIGIT SUM}
In order to test our approach on a more complex specification, we study the following task: Given a pair of MNIST digits, we ask the question: Can an attacker perturb each image, subject to a constraint on the total perturbation across both digits, such that the sum of the digits predicted by the classifier differs from the true sum of those digits by as large an amount as possible? Answering this question requires solving the following optimization problem:
\begin{align*}
 \max_{\substack{\xin_a, \xin_b\\ \epsilon_a, \epsilon_b}} & |\argmax\br{\post{L}\br{\xin_a}} + \argmax\br{\post{L}\br{\xin_b}}-s|    \\
 \text{s.t. } & \norm{\xin_a-\xnom_a} \leq \epsilon_a,  \norm{\xin_b-\xnom_b} \leq \epsilon_b\\
& \epsilon_a + \epsilon_b \leq \epsilon
\end{align*}
where $s$ is the true sum of the two digits. Thus, the adversary has to decide on both the perturbation to each digit, as well as the size of the perturbation. We can encode this within our framework (we skip the details here). The upper bound on the maximum error in the predicted sum from the verification and the lower bound on the maximum error computed from an attack for this problem (on an adversarially trained two hidden layer sigmoid network) is plotted in figure \ref{fig:complex_verify}. The results show that even on this rather complex verification task, our approach is able to compute tight bounds.

\section{CONCLUSIONS}
We have presented a novel framework for verification of neural networks. Our approach extends the applicability of verification algorithms to arbitrary feedforward networks with any architecture and activation function and to more general classes of input constraints than those considered previously (like cardinality constraints). The verification procedure is both efficient (given that it solves an unconstrained convex optimization problem) and practically scalable (given its anytime nature only required gradient like steps). We proved the first known (to the best of our knowledge) theorems showing that under special assumptions, nonlinear neural networks can be verified tractably. Numerical experiments demonstrate the practical performance of our approach on several classes of verification tasks.

\section*{ACKNOWLEDGEMENTS}
The authors would like to thank Brendan O'Donoghue, Csaba Szepesvari, Rudy Bunel, Jonathan Uesato and Shane Legg for helpful comments and feedback on this paper. Jonathan Uesato's help on adversarial training of neural networks is also gratefully acknowledged.


%% file: Appendix.tex
\section{APPENDIX}
\subsection{BOUND TIGHTENING}\label{sec:BoundTighten}
The quality of the bound in theorem \ref{thm:Main} depends crucially on having bounds on all the intermediate pre and post activations $\pre{l}, \post{l}$. In this section, we describe how these bounds may be derived. One simple way to compute bounds on neural activations is to use interval arithmetic given bounds on the input $\postl{0} \leq x \leq \postu{0}$. Bounds at each layer can then be computed recursively for $l=0, \ldots, L-1$ as follows:
\begin{subequations}
\begin{align}
&\prel{l} = \pos{\weight{l}}\postl{l} + \nega{\weight{l}}\postu{l} + \bias{l} \\
&\preu{l} = \pos{\weight{l}}\postu{l} + \nega{\weight{l}}\prel{l} + \bias{l} \\
&\postl{l+1} = \tf{l}\br{\prel{l}} \\
&\postu{l+1} = \tf{l}\br{\preu{l}}
\end{align}\label{eq:BoundProp}
\end{subequations}
However, these bounds could be quite loose and could be improved by solving the optimization problem
\begin{subequations}
\begin{align}
\max_{\substack{\pre{0}, \ldots, \pre{L-1}\\ \post{1}, \ldots, \post{L-1}}} & \post{l}_k
\\
 \text{s.t } & \eqref{eq:RobustVerifya}, \eqref{eq:RobustVerifyb}, \eqref{eq:RobustVerifyc}
\end{align} \label{eq:BT}
\end{subequations}
We can relax this problem using dual relaxation approach from the previous section and the optimal value of the relaxation would provide a new (possibly tighter) upper bound on $\post{l}_{k}$ than $\postu{l}_{k}$. Further, since our relaxation approach is anytime (ie for any choice of dual variables we obtain a valid bound), we can stop the computation at any time and use the resulting bounds. This is a significant advantage compared to previous approaches used in \citep{bunel2017piecewise}. Similarly, one can obtain tighter upper and lower bounds on $\post{l}$ for each value of $l, k$. Given these bounds, one can infer tighter bounds on $\pre{l}$ using \eqref{eq:BoundProp}.

Plugging these tightened bounds back into \eqref{eq:DualObj} and rerunning the dual optimization, we can compute a tighter upper bound on the verification objective.

\subsection{CONJUGATES OF TRANSFER FUNCTIONS}\label{sec:AppA}
\newcommand{\yl}{\underline{y}}
\newcommand{\yu}{\overline{y}}
We are interested in computing $g^\star = \max_{y \in [\yl, \yu]}g(y) = \mu y - \lambda \tfs\br{y}$. This is a one dimensional optimization problem and can be computed via brute-force discretization of the input domain in general. However, for most commonly used transfer functions, this can be computed analytically. We derive the analytical solution for various commonly used transfer functions:\\
\emph{ReLUs:} If $\tfs$ is a ReLU, $g(y)$ is piecewise linear and specifically is linear on $[\yl, 0]$ and on $[0, \yu]$ (assuming that $0 \in [\yl, \yu]$, else $g(y)$ is simply linear and can be optimized by setting $y$ to one of its bounds). On each linear piece, the optimum is attained at one of the endpoints of the input domain. Thus, the overall maximum can be obtained by evaluating $g$ at $\yl, \yu, 0$ (if $0 \in [\yl, \yu])$. Thus, \\ 
$g^\star = \begin{cases} \max\br{g(\yl), g(\yu), g(0)} & \text{ if } 0 \in [\yl, \yu] \\
\max\br{g(\yl), g(\yu)} & \text{ otherwise}\end{cases}$. \\
\emph{Sigmoid:} If $\tfs$ is a sigmoid, we can consider two cases: a) The optimum of $g$ is obtained at one of its input bounds or b) The optimum of $g$ is obtained at a point strictly inside the interval $[\yl, \yu]$. In case (b), we require that the derivative of $g$ vanishes, i.e:$\mu - \lambda\sigma(y) (1 - \sigma(y))=0$. Since $\sigma(y) \in [0, 1]$, this equation only has a solution if $\lambda \neq 0$ and 
$\frac{\mu}{\lambda} \in \left[0, \frac{1}{4}\right]$
In this case, the solutions are 
$\sigma(y) = \frac{1 \pm \sqrt{1 - \frac{4\mu}{\lambda}}}{2}$
Solving for $y$, we obtain
$y = \sigma^{-1}\br{\frac{1 \pm \sqrt{1 - \frac{4\mu}{\lambda}}}{2}}$
where $\sigma^{-1}$ is the logit function $\sigma^{-1}(t) = \log\br{\frac{t}{1-t}}$. We only consider these solutions if they lie within the domain $[\yl, \yu]$. Define:
\begin{align*}
y_1(\mu, \lambda) &= \max\br{\yl, \min\br{\yu, \frac{1 - \sqrt{1 - \frac{4\mu}{\lambda}}}{2}}}\\
y_2(\mu, \lambda) &= \max\br{\yl, \min\br{\yu, \frac{1 + \sqrt{1 - \frac{4\mu}{\lambda}}}{2}}}
\end{align*}

Thus, we obtain the following expression for $g^\star$:
\begin{align*}
&\begin{cases}
\max\br{g\br{\yl}, g\br{\yu}} \text{ if } \lambda = 0 \text{ or } \frac{\mu}{\lambda} \not\in [0,\frac{1}{4}] \\
\max\br{g\br{\yl}, g\br{\yu}, g\br{y_1(\mu,\lambda)}, g\br{y_2(\mu,\lambda)}} 
 \text{ otherwise }
\end{cases}
\end{align*}
\emph{Tanh:} Define 
\begin{align*}
& y_1(\mu,\lambda) =  \max\br{\yl, \min\br{\yu, \arctanh\br{\sqrt{1 - \frac{\mu}{\lambda}}}}}\\
& y_2(\mu,\lambda) = \max\br{\yl, \min\br{\yu, \arctanh\br{\sqrt{1 - \frac{\mu}{\lambda}}}}}
\end{align*}
and obtain $g^\star$ to be
\begin{align*}
&\begin{cases}
\max\br{g\br{\yl}, g\br{\yu}} \text{ if } \lambda = 0 \text{ or } \frac{\mu}{\lambda} \not\in [0,1] \\
\max\br{g\br{\yl}, g\br{\yu}, g\br{y_1(\mu,\lambda)}, g\br{y_2(\mu,\lambda)}} \\
 \text{ otherwise }
\end{cases}
\end{align*}
\subsubsection{MaxPool}\label{sec:AppMaxPool}
If $h$ is a max-pool, we need to deal with it layer-wise and not component-wise. We have $h\br{y}=\max\br{y_1, \ldots, y_t}$ and are interested in solving for 
\[g^\star = \max_{y \in [\yl, \yu]} \tran{\mu}y-\lambda h\br{y}\]
Note here that $\lambda$ is a scalar while $\mu$ is a vector. This can be solved by considering the case of each component of $y$ attaining the maximumum separately. We look at the case where $y_i$ attains the maximum below:
\begin{align*}
    \max_{y \in [\yl, \yu], y_i \geq y_j \, \forall j \neq i}\tran{\mu}y - \lambda y_i
\end{align*}
Fixing $y_i$ and optimizing the other coordinates, we obtain
\begin{align*}
    \max_{y_i \in [\yl_i, \yu_i]}& \sum_{j \neq i, y_i \geq \yu_j}\max\br{\mu_j \yu_j, \mu_j\yl_j} \\
    & + \sum_{j \neq i, y_i \leq \yu_j} \max\br{\mu_j y_i, \mu_j\yl_j}\\
    & - \br{\mu_i - \lambda} y_i
\end{align*}
This one-dimensional function can be optimized via binary search on $y_i$. After solving for each $i$, taking the maximum over $i$ gives the value 
$g^\star$
\subsubsection{Upper bounds for general nonlinearities}\label{sec:AppAGeneralConj}
In general, we can compute an upper bound on $g$ even if we cannot optimize it exactly. The idea is to decouple the $y$ from the two terms in $g$:
$\mu y$ and $-\lambda h(y)$ and optimize each independently. However, this gives a very weak bound. This can be made much tighter by applying it separately to a decomposition of the input domain $[\yl, \yu]=\cup_i [a_i, b_i]$:
\[\max_{y \in [a_i, b_i]} g(y) \leq \max(\mu a_i, \mu b_i) + \max\br{-\lambda h(a_i), -\lambda h(b_i)}\]
Finally we can bound $\max_{[\yl, \yu]} g(y)$ using
\begin{align*}
& \max_{y \in [\yl, \yu]} g(y) \leq \\
& \max_i \br{\max(\mu a_i, \mu b_i) + \max\br{-\lambda h(a_i), -\lambda h(b_i)}}    
\end{align*}
As the decomposition gets finer, ie, $|a_i - b_i| \to 0$, we obtain an arbitrarily tight upper bound on $g$) this way.
\subsection{OPTIMIZING OVER THE INPUT CONSTRAINTS}\label{sec:AppB}
In this section, we discuss solving the optimization problem defining $f_0(\mum{0})$ in \eqref{eq:DualOpt}
\[\max_{x \in \Sin} (W^T\mu)^T x + b^T\mu\]
where we dropped the superscript $(0)$ for brevity. For commonly occuring constraint sets $\Sin$, this problem can be solved in closed form easily:\\
\emph{Norm constraints}: Consider the case $\Sin=\norm{x-\hat{x}} \leq \epsilon$. In this case, we by Holder's inequality, the objective is larger than or equal to
\[b^T\mu - \norm{W^T\mu}_\star\]
where $\norm{\cdot}_\star$ is the dual norm to the norm $\norm{}$. Further this bound can be achieved for an appropriate choice of $x$. Hence, the optimal value is precisely $b^T\mu - \norm{W^T\mu}_\star$. \\
\emph{Combinatorial objects}: Linear objectives can be optimized efficiently over several combinatorial structures. For example, if $x$ is indexed by the edges in a graph and $\Sin$ imposes constraints that $x$ is binary valued and that the edges set to $1$ should form a spanning tree of the graph, the optimal value can be computed using a maximum spanning tree approach. \\
\emph{Cardinality constraints}: $x$ may have cardinality constrained imposed on it: $\norm{x}_0\leq k$, saying that at most $k$ elements of $x$ can be non-zero. If further we have bounds on $x \in [\underline{x}, \overline{x}]$, then the optimization problem can be solved as follows: Let $v(\mu) = \pos{W^T \mu} \odot \overline{x} + \nega{W^T \mu} \odot \underline{x}$ and let $[v]_i$ denote the $i$-th largest component of $v$. Then the optimal value is $\sum_{i=1}^k [v(\mu)]_i + b^T \mu$.

\subsection{PROOFS OF THEORETICAL RESULTS}\label{sec:TheoryRes}
\subsubsection{NP-hardness of a verification of a single hidden layer network}\label{sec:NP-hard}
Consider the case of sigmoid transfer function with an $\infty$ norm perturbation. Then, the verification problem reduces to :
\begin{align*}
\max_{\xin, z} & \sum_i c_i \tfs_i\br{z_i} \\
\text{ s.t } & z= Wx+b, \norm{\xx-\xnom}_\infty \leq \epsilon
\end{align*}
This is an instance of a sigmoidal programming problem, which is proved to be NP-hard in \citet{udell2013maximizing}

\subsubsection{Proof of theorem \ref{thm:Planet}}\label{sec:AppCompareRelax}
\begin{proof}
In this case, since $h$ is a relu, $\tilde{f}$ can be written as (from section \ref{sec:AppA}) as 
\begin{align}
    \tilde{f}_{l, k}(\lambda, \mu) = \begin{cases}
    \max\br{\mu \prel{l}_k, \br{\mu-\lambda} \preu{l}_k, 0} \text{ if } 0 \in [\prel{l}_k, \preu{l}_k]\\
    \max\br{\mu \preu{l}_k, \mu \prel{l}_k}  \text{ if } 0 \text{ if } \preu{l}_k \leq 0\\
    \max\br{\br{\mu-\lambda} \prel{l}_k, \br{\mu-\lambda} \preu{l}_k}  \text{ if } \prel{l} \geq 0
\end{cases}\label{eq:zcases}
\end{align}
Now, the LP relaxation from \citep{ElhersPlanet} can be written as
\begin{align*}
& \max c^T \post{L} \\
& \text{s.t } \pre{l}=\weight{l}\post{l}+\bias{l}, l=0, \ldots, L-1 \\
& \post{l+1}_k = \pre{l}_k  \quad \forall l, k \text{ s.t } \prel{l}_k \geq 0\\
& \post{l+1}_k = 0 \text{ if }  \quad \forall l, k \text{ s.t } \preu{l}_k \leq 0\\
& \post{l+1}_k \geq 0, \post{l+1}_k \geq \pre{l}_{k}, \post{l+1}_k \leq \br{\pre{l}_k-\prel{l}_k}\br{\frac{\preu{l}_k}{\preu{l}_k-\prel{l}_k}} \\
& \qquad \quad \quad \forall l, k \text{ s.t } 0 \in [\prel{l}_k, \preu{l}_k]\\
& \qquad \prel{l} \leq \pre{l} \leq \preu{l}, \postl{l} \leq \post{l} \leq \postu{l} \quad \forall l
\end{align*}
We can rewrite this optimization problem as  
\begin{align*}
& \max c^T \post{L} \\
& \text{s.t } \pre{l}=\weight{l}\post{l}+\bias{l}, l=0, \ldots, L-1 \\
& \post{l+1}_k = \pre{l}_k  \quad \forall l, k \text{ s.t } \prel{l}_k \geq 0\\
& \post{l+1}_k = 0 \text{ if }  \quad \forall l, k \text{ s.t } \preu{l}_k \leq 0\\
& \post{l+1}_k \geq \max\br{0, \pre{l}_{k}}, \post{l+1}_k \leq \br{\pre{l}_k-\prel{l}_k}\br{\frac{\preu{l}_k}{\preu{l}_k-\prel{l}_k}} \\
& \qquad \quad \quad \forall l, k \text{ s.t } 0 \in [\prel{l}_k, \preu{l}_k]\\
& \qquad \prel{l} \leq \pre{l} \leq \preu{l}, \postl{l} \leq \post{l} \leq \postu{l} \quad \forall l
\end{align*}
which is still a convex optimization problem, since all the constraints are either linear of the form $\max(0, z) \leq x$ which is a  convex constraint since the LHS is a convex function and the RHS is linear.

Taking the dual of this optimization problem, we obtain
\begin{align*}
& \max_{x, z} c^T \post{L} + \sum_l \tran{\mum{l}}\br{\pre{l}-\weight{l}\post{l}-\bias{l}}\\
& \quad +\sum_{l, k: \prel{l}_k \geq 0} \lam{l}_k\br{\post{l+1}_k-\pre{l}_k} \\
& \quad +\sum_{l, k: \prel{l}_k \leq 0} \lam{l}_k\br{\post{l+1}_k} \\
& \quad +\sum_{l, k: 0 \in [\prel{l}_k, \preu{l}_k]} \br{\lam{l}_{k;a}\br{\post{l+1}_k-\max\br{\pre{l}_k, 0}}}\\
& \quad +\sum_{l, k: 0 \in [\prel{l}_k, \preu{l}_k]}-\br{\post{l+1}_k-s^{l}_k\br{\pre{l}_k-\prel{l}_k}}\lam{l}_{k;b} \\
& \text{s.t } \qquad \prel{l} \leq \pre{l} \leq \preu{l}, \postl{l} \leq \post{l} \leq \postu{l} \quad \forall l 
\end{align*}
where $s^l_k=\frac{\preu{l}_k}{\preu{l}_k-\prel{l}_k}$. Let $\lam{l}_k = \lam{l}_{k;a}-\lam{l}_{k;b}$ for $l, k$ such that 
$0 \in [\prel{l}_k, \preu{l}_k]$. Further let $\mathcal{I}_{l, k} = [\prel{l}_k, \preu{l}_k]$ and let $\mathcal{I}_a$ denote the set of $l, k$ such that $\prel{l}_k \geq 0$, $\mathcal{I}_b$ the set of $l, k$ such that $\preu{l}_k \leq 0$ and $\mathcal{I}_c$ the set of $l, k$ such that $0 \in \mathcal{I}_{l,k}$. 
We can then rewrite the dual as
\begin{align*}
& \max_{x} c^T \post{L} + \sum_{l=0}^{L-1} \tran{\post{l}}\br{\lam{l}-\tran{\weight{l}}\mum{l}} -\sum_l \tran{\mum{l}}\bias{l} \\
& + \sum_{l, k \in \mathcal{I}_a} \max_{\pre{l}_k \in [\prel{l}_k, \preu{l}_k]} \br{\mum{l}_k-\lam{l}_k} \pre{l}_k \\ & + \sum_{l, k \in \mathcal{I}_b} \max_{\pre{l}_k \in [\prel{l}_k, \preu{l}_k]} \br{\mum{l}_k} \pre{l}_k \\ & + \sum_{l, k \in \mathcal{I}_c} \max_{\pre{l}_k \in \mathcal{I}_{l,k}} \br{\mum{l}_k+\lam{l}_{k;b}s^l_k} \pre{l}_k-\br{\lam{l}_{k;a}}\max\br{\pre{l}_k, 0}\\ & + \sum_{l, k \in \mathcal{I}_c}-\lam{l}_{k,b}s^l_k\prel{l}_k \\
& \text{s.t } \postl{l} \leq \post{l} \leq \postu{l} \quad \forall l 
\end{align*}

We now solve for the maximum over $\pre{l}_k$ considering three cases: \\
(a) $l, k \in \mathcal{I}_a$: The maximization is over a linear function and the maximum is attained at one of the bounds, hence the maximum evaluates to
\[\max\br{\br{\mum{l}_k-\lam{l}_k}\prel{l}_k, \br{\mum{l}_k-\lam{l}_k}\preu{l}_k}\]
This evaluates to $\tilde{f}_{l, k}\br{\lam{l}_k, \mum{l}_k}$ for $l, k  \in \mathcal{I}_a$.
(b) $l, k \in \mathcal{I}_b$: The maximization is over a linear function and the maximum is attained at one of the bounds, hence the maximum evaluates to
\[\max\br{\br{\mum{l}_k}\prel{l}_k, \br{\mum{l}_k}\preu{l}_k}\]
This evaluates to $\tilde{f}_{l, k}\br{\lam{l}_k, \mum{l}_k}$ for $l, k  \in \mathcal{I}_b$.
(c) $l, k \in \mathcal{I}_b$: The maximization is over a piecewise linear function and the maximum is attained at one of the bounds or at the breakpoint $0$, hence the maximum evaluates to
\[\max\br{0, \br{\mum{l}_{k} + \lam{l}_{k,b}s^l_k}\prel{l}_k, \br{\mum{l}_{k} + \lam{l}_{k,b}\br{s^l_k-1}-\lam{l}_k }\preu{l}_k}\]
Adding the constant $-\lam{l}_{k,b}s^l_k\prel{l}_k$ we obtain
\[\max\br{-\lam{l}_{k,b}s^l_k\prel{l}_k, \mum{l}_{k}\prel{l}_k, \br{\mum{l}_{k} -\lam{l}_k }\preu{l}_k}\]
Minimizing the above expression with respect to $\lam{l}_{k;b}$ subject to $\lam{l}_{k;b}\geq \max\br{0, -\lam{l}_k}$, we obtain
\[\max\br{0, \lam{l}_k\frac{\preu{l}_k\prel{l}_k}{\preu{l}_k-\prel{l}_k}, \mum{l}_{k}\prel{l}_k, \br{\mum{l}_{k} -\lam{l}_k }\preu{l}_k}\]
(since the expression is monotonically increasing in $\lam{l}_{k;b}$, minimizing it subject to these constraints we just set $\lam{l}_{k;b}$ to the larger of its lower bounds). Now, for the second term to attain the maximum, we require that $\lam{l}_k \leq 0, \mum{l}_k = s^l_k \lam{l}_k$, in which case all the last three terms attain the maximum. Thus, the maximum is equal to the max of three terms:
\[\max\br{0, \mum{l}_{k}\prel{l}_k, \br{\mum{l}_{k} -\lam{l}_k }\preu{l}_k}\]
showing that the expression evaluts to $\tilde{f}_{l, k}\br{\lam{l}_k, \mum{l}_k}$ for $l, k \in \mathcal{I}_c$.

Thus, the dual objective is equal to 
\begin{align*}&\max_{x: \postl{l} \leq \post{l} \leq \postu{l}} c^T \post{L} + \sum_{l=0}^{L-1} \tran{\post{l}}\br{\lam{l}-\tran{\weight{l}}\mum{l}}\\
 & -\sum_l \tran{\mum{l}}\bias{l} \\
& + \sum_{l, k} \tilde{f}_{l, k}\br{\lam{l}_k, \mum{l}_k}\end{align*}

 Given this, the rest of the dual exactly matches the calculations from section \ref{sec:MainDual}. 

\end{proof}

\subsubsection{Proof of theorem \ref{thm:Exact}}
\begin{proof}
We leverage results from \citep{polyak2003convexity} which argues that a smooth nonlinear function can be efficicently optimized over a ``small enough'' ball. Specifically, we use theorem 7 from \citep{polyak2003convexity}. In order to apply the theorem, we need to bound the Lipschitz costant of the derivative of the function $f(x)=\sum_i c_i h_i(W_ix+b_i)$. Writing down the derivative, we obtain
\[f^\prime(x) = W^T \diag\br{c}h^\prime(Wx+b)\]
Thus,
\begin{align*}
& f^\prime(x) - f^\prime(y) \\
& = W^T \diag\br{c}\br{h^\prime(Wx+b)-h^\prime(Wy+b)}
\end{align*}
We have 
\[h^\prime_i(W_ix+b_i)-h^\prime_i(W_iy+b_i)=h_i^{\prime\prime}\br{t}(W_i(x-y))\] 
where $t \in [W_ix+b_i, W_iy+b_i]$. Thus, we have 
\[|h^\prime_i(W_ix+b_i)-h^\prime_i(W_iy+b_i)| \leq \gamma_i |W_i(x-y)|^2\]
So that 
\begin{align*}
& \norm{f^\prime(x) - f^\prime(y)}_2  \\
& \leq \sigma_{\max}\br{W^T \diag\br{c}}\norm{h^\prime(Wx+b)-h^\prime(Wy+b)}_2\\
& = \sigma_{\max}\br{W^T \diag\br{c}}\times \\
& \qquad \sqrt{\sum_i {\br{h^\prime_i(W_ix+b_i)-h^\prime_i(W_iy+b_i)}}^2} \\
& \leq \sigma_{\max}\br{W^T \diag\br{c}}\gamma \sqrt{\sum_i |\gamma_i W_i(x-y)|^2}  \\
& = \sigma_{\max}\br{W^T \diag\br{c}}\norm{\diag\br{\gamma}W(x-y)}_2\gamma \\
& \leq \sigma_{\max}\br{W^T \diag\br{c}}\sigma_{\max}\br{\diag\br{\gamma}W}\norm{x-y}_2
\end{align*}
Thus $f^\prime$ is Lipschitz with Lipschitz constant $\sigma_{\max}\br{W^T \diag\br{c}}\sigma_{\max}\br{W}\gamma$. Further
\[\norm{f^\prime\br{x^0}} = \norm{W^T\diag\br{c}h^\prime\br{Wx^0+b}}\]
Hence by theorem 7 from \citep{polyak2003convexity}, the theorem follows.
\end{proof}

\subsubsection{PROOF OF THEOREM \ref{thm:Approx}}

\begin{proof}
We have $\norm{z}_2\leq \epsilon \implies |W_iz| \leq \norm{W_i}_2\epsilon$ (by Cauchy-Schwartz). Thus, for eacn $i$, we have
\begin{align*}
& h_i(\znom_i+W_iz) =   \\
& =h_i(\znom_i)+h_i^\prime(\znom_i)W_iz+\frac{h_i^{\prime\prime}(\znom_i)}{2}\br{W_iz}^2 \\
& \, +\frac{1}{6} h_i^{\prime\prime\prime}(\tilde{x}^0_i+t_i\br{W_iz})\br{W_iz}^3
\end{align*}
The last term can be bounded above by 
\[\frac{\eta_i}{6}|W_iz|^3 \leq \frac{\eta_i}{6}\zeta_i^3\epsilon^3\]
Adding the error terms over all terms in the objective function, we obtain the result.
\end{proof}